\documentclass[]{article}
\usepackage{lmodern}
\usepackage{amssymb,amsmath}
\usepackage{ifxetex,ifluatex}
\usepackage{fixltx2e} 
\ifnum 0\ifxetex 1\fi\ifluatex 1\fi=0 
  \usepackage[T1]{fontenc}
  \usepackage[utf8]{inputenc}
\else 
  \ifxetex
    \usepackage{mathspec}
    \usepackage{xltxtra,xunicode}
  \else
    \usepackage{fontspec}
  \fi
  \defaultfontfeatures{Mapping=tex-text,Scale=MatchLowercase}
  
\fi
\IfFileExists{upquote.sty}{\usepackage{upquote}}{}
\IfFileExists{microtype.sty}{\usepackage{microtype}}{}
\ifxetex
  \usepackage[setpagesize=false, 
              unicode=false, 
              xetex]{hyperref}
\else
  \usepackage[unicode=true]{hyperref}
\fi
\hypersetup{breaklinks=true,
            bookmarks=true,
            pdfauthor={Phillip M. Alday},
            pdftitle={Be Careful When Assuming the Obvious: Commentary on The placement of the head that minimizes online memory: a complex systems approach},
            colorlinks=true,
            citecolor=blue,
            urlcolor=blue,
            linkcolor=magenta,
            pdfborder={0 0 0}}
\title{Be Careful When Assuming the Obvious: Commentary on ``The placement of
the head that minimizes online memory: a complex systems approach''}
\author{Phillip M. Alday}

\begin{document}
\maketitle

\begin{abstract}
Ferrer-i-Cancho (in press) presents a mathematical model of both the synchronic and diachronic nature of word order based on the assumption that memory costs are a never decreasing function of distance and a few very general linguistic assumptions. 
However, even these minimal and seemingly obvious assumptions are not as safe as they appear in light of recent typological and psycholinguistic evidence. 
The interaction of word order and memory has further depths to be explored.
\end{abstract}


\emph{Sprachwissenschaft? Gauß wiegte den Kopf. Das sei etwas für Leute,
welche die Pedanterie zur Mathematik hätten, nicht jedoch die
Intelligenz. Leute, die sich ihre eigene notdürftige Logik erfänden.}
(Kehlmann 2007, 169)\footnote{Haider (2009) provides a good translation
  of the fictional dialog between Gauss and Humboldt: ``Linguistics?
  This is something for people with the pedantry, but not the
  intelligence, for mathematics. People who invent their own scanty
  logics.''}

Linguistics, especially in its more cognitively oriented forms, has an
uneasy relationship with mathematics. Many of the formalisms used since
the Chomskyan revolution --- trees, transformations, aspects of
combinatorics, etc. --- are clearly mathematical in nature, yet their
use in linguistics is decidedly less rigorous than their use in the
computational sciences, e.g.~the study of automata. These parallels are
perhaps responsible for the disturbing trend to move further away from
data towards an idealized platonic world of language. Worse, the lack of
rigor means that the theories have become increasingly baroque, yet
still unable to explain core phenomena. Ferrer-i-Cancho (in press) shows
the weakness of this approach by addressing a long-standing curiosity of
linguistic typology, namely the preference for certain word order
configurations both synchronically and diacronically, with an elegant
and theory-agnostic approach. But just as the obvious truths of
Euclidean geometry require some further consideration in our decidedly
non-Euclidean universe, Ferrer-i-Cancho's axioms must be considered
carefully in light of the complexity of memory and language.

Ferrer-i-Cancho's approach is remarkable for a number of reasons. In
terms of linguistic theory, the primary assumption is that there is
something like a head-dependent relationship in a very broad form --- so
broad indeed that the definition would apply equally well to both
phrase-structure and dependency grammars. While the algebra involved is
occasionally tedious, the proofs do not use any particularly advanced
machinery.\footnote{I think Ferrer-i-Cancho would agree with me that the
  fear of equations even at the level in his paper is one of the bigger
  problems in moving the study of language away from its metaphysical
  beginnings towards a more empirical study. Haider (2009) seems also to
  follow this spirit, in discussing the necessity of moving from
  very-well described data to (mathematical) characterization of that
  data (basic models expressing the nature of certain relationships,e.g.
  \emph{laws} in the physical sciences) and further to explanations of
  those relationships which make predictions beyond the initial data.
  Ferrer-i-Cancho's contribution seems to be the first step in moving
  from description to prediction.}

The assumptions regarding memory are also relatively simple: the more
intervening elements between a head and its dependents, the higher the
memory cost; more specifically, the cost always increases with each
additional element. For simplicity's sake, Ferrer-i-Cancho also assumes
that the relationship is fully symmetric --- it does not matter if the
dependent comes before the head or vice versa. Other typological
features --- morphological alignment, synthetic vs.~analytic --- are not
impacted by this generality, and this is a strength. Ferrer-i-Cancho
does not remark upon this, but the difficulties in defining terms like
``subject'' from a cross-linguistic perspective are not a problem here
--- although the labels ``S'' and ``O'' are used, the results hold for
the arguments of any transitive relationship, regardless of how
agreement and case-marking work in the language in question.\footnote{Traditional
  grammatical notions of ``subject'' (tied to the nominative case,
  agreement and word order) quickly fall apart when we move beyond
  western Europe. Ergativity presents an obvious difficulty as does
  topic prominence, e.g.~in Chinese (LaPolla 1993).} Moreover, although
Ferrer-i-Cancho focuses on the configuration of S, V and O, his core
assumptions would also be valid for any configuration involving a head
and two dependents. This would explain, for example, why conjunctions
often appear between conjoined elements, although e.g.~a suffix on the
second element (cf.~Latin \emph{-que}) would also be possible. The
generality and power of this result follow directly from the minimal
assumptions.

However, this generality presents a problem in and of itself. The
stability of head-central placement should also hold for other head
types, yet there are no languages which place half of the adjectives
before the head noun and the other half after it. (And if morphology
uses the same machinery as syntax, then the relative rarity of infixes
compared to prefixes and suffixes would also pose a problem.) Perhaps
this can be explained by the optionality of further dependents. If nouns
most commonly occur with only one or no modifier, then central placement
would be a special case. (The lack of noun phrases with many modifiers
is probably also related to certain aspects of online memory cost.) The
simpler processing mechanism is then to pick left- or right-headedness
and stick with it. In computational terms, special cases aren't special
enough --- the computational cost of a single mechanism is less than the
memory cost incurred by not using special cases, when amortized over all
sentences.

Ferrer-i-Cancho posits that the (anti-)symmetry of head placement in NPs
compared to the verbal head at the sentence level may serve as a
counterweight in preserving the stability of certain word order
configurations. This explains why the central placement usually does not
occur, but the question that remains is why there are \emph{no}
languages with central noun placement. Interestingly, in his derivation
of this result, he also provides a more rigorous explanation for
consistent headedness across phrase types --- head-right for NPs (and,
by extension, postpositions instead of prepositions) actually reduces
the total dependency length summed over multiple structural levels for
SOV word order, while there is no clear advantage for pre- nor
postnominal modifiers in the symmetric world of SVO, which might explain
the post-nominal adjective placement with \emph{pre}positions observed
in the Romance languages. (Indeed, the mixture of prepositions and
post-nominal modifiers could be seen as an attempt to place the nominal
head centrally.) The deeper yet unexplored implication here is an
explanation for certain regularities in branching direction across
phrase types that is much more satisfying --- and better able to deal
with certain irregularities --- than the Chomskyan assumption of a
single parameter for the relative placements of heads.

The three-element representation of word order is a simplification that
struggles even with the Indoeuropean languages. Verb-second word order
in the Germanic languages (cf. Mallinson and Blake 1981; Haider 2010)
admits SVO and OVS as some of its possibilities; however, VSO and VOS
can also appear under certain conditions. In subordinate clauses, the
Germanic languages often show yet a different word order, and the
subtleties are made even worse by the (asymmetric) verb-second property
of German: in unmarked declarative clauses, the finite verb appears in
the left periphery of the clause (the so-called ``left sentence
bracket'', in terms of traditional grammatic description), preceded by a
single constituent, whereas in canonical embedded clauses, it usually
stays in final position (``right sentence bracket'', Harbert 2007). (In
both cases, the infinite part of the verb appears canonically in the
right periphery.) In Italian, the presumed word order is SVO (Bates et
al. 1982; cf.~the entry for Italian in the World Atlas of Structures,
Dryer and Haspelmath 2013), yet object pronouns appear as preverbal
clitics and subject pronouns are often omitted. Full NP objects can
appear initially and full NP subjects can appear post-verbally under
certain information-structural conditions (Bianchi; cf.~the behavioral
data in Bates et al. 1982). Synchronic word order is unfortunately not
as static as it appears. On the other hand, Ferrer-i-Cancho clearly
states that he is modelling \textbf{only} the role of dependency length
in determining word order and not any other factors. From a cognitive
perspective, however, the question remains to be answered whether memory
costs are the decisive factor in determining an optimal word order.

In speaking about the optimality of the verbal encoding of language, we
also need to consider that what is optimal for the speaker may not be
optimal for the listener. Because the majority of research into the role
of memory in language processing deals with language
perception\footnote{There are perhaps many reasons for this, but one of
  the clearest is that it is far easier to perform controlled perception
  experiments.}, and the notion of ``open dependencies'' seems to match
up best with the notion of incremental processing during language
perception, I will focus on some issues with the assumptions about
memory with respect to perception. In psycholinguistics, Gibson
formulated his Syntactic Prediction Locality Theory (SPLT, 1998) and its
evolution, the Dependency Locality Theory (DLT, 2000), on the basis of
empirical evidence that processing difficulty increased with the number
of elements between dependent entities. Indeed, Ferrer-i-Cancho and
Gibson make many of the same assumptions, although Gibson worked more
with processing difficulty of individual sentences and less with
typological trends. Gibson's theory is able to explain a great many
effects reported in the psycholinguistic literature; however,
anti-locality effects, where a more distant element is processed
\emph{more} quickly, present a difficult problem for such models
(Konieczny 2000; Lewis, Vasishth, and Dyke 2006).

Recent models of short-term memory (cf. Jonides et al. 2008) posit a
content-addressable system capable of direct access. More precisely,
these models have feature-based indexing with some notion of rapid
decay, corresponding to notions of ``focus'' in traditional memory
models (which often divide memory into focus, short- and long-term
memory). In such models, distance does play a role through an initially
very rapid decay in activation followed by an asymptotic dwindling, but
because of the rapid nature of the initial decay in activation, distance
is generally not the dominating factor. Rather, the additional
intervening elements often have overlapping features, which lead to
interference, which increases memory costs. Because of the way
activation is implemented in the mind / brain, having additional
intervening elements does not cost more in and of itself, but rather
increases the costs of retrieving any particular previous element
(e.g.~the head), potentially to the point of being unaffordable
(i.e.~forgetting). In terms of computation, we can think of
dependency-length models as being like a search through an array (or
even a stack, when we consider it in context of the classic Sternberg
task), while newer feature-based models are comparable to hash-based
lookup, where overlapping features lead to hash-collisions. Lewis and
colleagues (Lewis and Vasishth 2005; Lewis, Vasishth, and Dyke 2006)
applied this type of memory model to language processing and showed that
it could explain the dependency-length effects in the literature
\emph{and} the anti-locality effects.

However, sequence-related information appears to be stored as a partial
ordering, necessitating a serial search of sorts, so there is indeed
some aspect of memory involving linear distance.\footnote{Even in the
  idealized world of computer science, hash collisions can lead to
  linear search.} Summarizing previous work, McElree, Foraker, and Dyer
(2003) posit that the content-addressable memory subsystem is
complemented by one using serial search for ``temporal order'' and
``position''. Bornkessel-Schlesewsky and Schlesewsky (2013) distinguish
between ``time-dependent'' and ``time-independent'' computations in
language processing, i.e.~commutative and non-commutative operations.

This complexities of addressing for storage and retrieval tie into a
somewhat deeper issue regarding the nature of memory in language
processing. Although newer memory models posit a feature-based recall
system with falling activation, this does not mean that individual
sentence entities are stored in the received form. Due to the
incremental and predictive nature of language processing, it is quite
possible that a form of partial evaluation takes place, e.g.~combining
separate noun and verb entities into one Action entity or combining a
subject and an object into one Relation (e.g.~Cause-Effect) entity (cf.
\emph{Actor} in Bornkessel-Schlesewsky and Schlesewsky 2009; Bornkessel
and Schlesewsky 2006; but older notions of purely syntactic folding or
``chunking'' can be traced back to the Sausage Machine in Frazier and
Fodor 1978 and classical parsing strategies like Minimal Attachment and
Late Closure). In this vein, the distance to the head may be less
relevant than distance to the next element in the relation, even when
that element is a fellow dependent.

Consider for example a typical transitive construction in an SOV
language --- many aspects of the relationship between S and O will
already be established at the SO position,\footnote{Ferrer-i-Cancho
  references Lupyan and Christiansen in considering that case marking
  may facilitate the learning of SOV structures. This also fits quite
  well with incremental relational processing --- case marking
  establishes the relationship between the nominal arguments even in the
  (temporary) absence of a verb (cf. Bornkessel-Schlesewsky and
  Schlesewsky 2009).} which means that the S may not need to be bound to
the verb, but rather to its sister dependency O and that the distance
between this conjoined dependency and its head verb is decisive. In this
case, the summed distance from S to O and from O to V is more important
than the summed SV and OV distances. Moreover, in some extreme cases,
the strength of the prediction may be so strong that not all elements
are fully processed, as seen for example in the Moses illusion (A. J.
Sanford et al. 2010). Alday, Schlesewsky, and Bornkessel-Schlesewsky
(2014) were able to model neurolinguistic data by implementing this type
of relational processing via a form of weighted feature overlap for
features prototypical to the roles in transitive relations. Even for a
strictly dependency-length based model, such chunking may reduce the
length of the dependency by folding the intervening elements together,
which would violate the assumption of strict monotonicity (of the cost
of a dependency as a function of its length that is at the basis of
Ferrer-i-Cancho's argument).

In spite of the challenges presented above, Ferrer-i-Cancho's model is a
step in the right direction. He presents a general yet rigorous
explanation for both synchronic and diachronic tendencies in word order
across languages and thus formalizes a long-standing intuition amongst
more cognitively oriented linguists. Moreover, he shows the importance
in exploring the role of simple mechanisms in complex systems like
language. However, like all such mathematical models, the simplicity and
elegance come at a certain cost, namely obscuring many of the messy
details of reality. Euclidean geometry is a good-enough approximation
for many everyday phenomena but its ultimately flawed assumptions are
insufficient for a truly deep understanding of our curved space-time. In
much the same way, Ferrer-i-Cancho's model is a good starting point,
especially for the big picture of diachronic change, but cannot be the
last stop if we are to fully understand the cognitive mechanisms of word
order.

\section{Acknowledgement}\label{acknowledgement}

I would like to thank R. Ferrer-i-Cancho for the interesting discussion
and essential feedback.

\section{Bibliography}\label{bibliography}

Alday, Phillip M., Matthias Schlesewsky, and Ina Bornkessel-Schlesewsky.
2014. ``Towards a Computational Model of Actor-Based Language
Comprehension.'' \emph{Neuroinformatics} 12 (1): 143--179.

Bates, Elizabeth, Sandra McNew, Brian MacWhinney, Antonella Devescovi,
and Stan Smith. 1982. ``Functional Constraints on Sentence Processing: A
Cross-Linguistic Study.'' \emph{Cognition} 11: 245--299.

Bianchi, Valentina. ``On Focus Movement in Italian.'' In
\emph{Information Structure and Agreement}, edited by M. V.
Camacho-Taboada, A. Jiménez Fernández, and Reyes-Tejedor Martín-Gonzáles
J. Amsterdam \& Philadelphia: John Benjamins.

Bornkessel, Ina, and Matthias Schlesewsky. 2006. ``The Extended Argument
Dependency Model: A Neurocognitive Approach to Sentence Comprehension
Across Languages.'' \emph{Psychological Review} 113 (4): 787--821.

Bornkessel-Schlesewsky, Ina, and Matthias Schlesewsky. 2009. ``The Role
of Prominence Information in the Real-Time Comprehension of Transitive
Constructions: A Cross-Linguistic Approach.'' \emph{Language and
Linguistics Compass} 3 (1): 19--58.

---------. 2013. ``Reconciling Time, Space and Function: A New
Dorsal-Ventral Stream Model of Sentence Comprehension.'' \emph{Brain and
Language} 125 (1): 60--76.

Dryer, Matthew S., and Martin Haspelmath, ed. 2013. \emph{WALS Online}.
Leipzig: Max Planck Institute for Evolutionary Anthropology.
\url{http://wals.info/}.

Ferrer-i-Cancho, Ramon. in press. ``The Placement of the Head That
Minimizes Online Memory: a Complex Systems Approach.'' \emph{Language
Dynamics and Change}.

Frazier, Lyn, and Janet Dean Fodor. 1978. ``The Sausage Machine: A New
Two-Stage Parsing Model.'' \emph{Cognition} 6 (4): 291--325.

Gibson, Edward. 1998. ``Linguistic Complexity: Locality of Syntactic
Dependencies.'' \emph{Cognition} (Jan).

---------. 2000. ``The Dependency Locality Theory: A Distance-Based
Theory of Linguistic Complexity.'' In \emph{Image, Language, Brain},
edited by Y Miyashita, A Marantz, and O'NeilW, 95--126. Cambridge, MA:
MIT Press.

Haider, Hubert. 2009. ``Linguistics -- Postmodern Frippery or Primordial
Cognitive Science?'' \emph{Presentation at the ``Visions for
Linguistics'' Workshop}. Schloss Freudental 20--22 Nov. 2009:
Universität Konstanz.

---------. 2010. \emph{The Syntax of German}. Cambridge: Cambridge
University Press.

Harbert, Wayne. 2007. \emph{The Germanic Languages}. Cambridge:
Cambridge University Press.

Jonides, J, R Lewis, D Nee, and C Lustig. 2008. ``The Mind and Brain of
Short-Term Memory.'' \emph{Annual Reviews} (Jan).

Kehlmann, Daniel. 2007. \emph{Die Vermessung Der Welt}. Hamburg:
Rowohlt.

Konieczny, Lars. 2000. ``Locality and Parsing Complexity.''
\emph{Journal of Psycholinguistic Research} 29 (6): 627--645.

LaPolla, Randy J. 1993. ``Arguments Against 'Subject' and 'Direct
Object' as Viable Concepts in Chinese.'' \emph{Bulletin of the Institute
of History and Philology} 63 (4): 759--813.

Lewis, R, and S Vasishth. 2005. ``An Activation-Based Model of Sentence
Processing as Skilled Memory Retrieval.'' \emph{Cognitive Science: A
Multidisciplinary Journal} (Jan): 375--419.

Lewis, R, S Vasishth, and J Van Dyke. 2006. ``Computational Principles
of Working Memory in Sentence Comprehension.'' \emph{Trends in Cognitive
Sciences} 10 (10) (Jan): 447--454.

Lupyan, G., and M. H. Christiansen. ``Case, Word Order, and Language
Learnability: Insights from Connection Modeling.'' In \emph{Proceedings
of the 24th Annual Conference of the Cognitive Science Society},
591--601. Mahwah, NJ: Lawrence Erlbaum.

Mallinson, Graham, and Barry J Blake. 1981. \emph{Language Typology :
cross-Linguistic Studies in Syntax}. Amsterdam: North-Holland
Publishers.

McElree, Brian, Stephani Foraker, and Lisbeth Dyer. 2003. ``Memory
Structures That Subserve Sentence Comprehension.'' \emph{Journal of
Memory and Language} 48 (1): 67--91.

Sanford, Anthony J, Hartmut Leuthold, Jason Bohan, and Alison J S
Sanford. 2010. ``Anomalies at the Borderline of Awareness: An ERP
Study.'' \emph{Journal of Cognitive Neuroscience} 23 (3): 514--523.

\end{document}